\pdfoutput=1
\documentclass{sig-alternate-05-2015}
\usepackage{multirow}
\usepackage{paralist}

\DeclareMathOperator*{\argmax}{arg\,max}
\DeclareMathOperator*{\ndcl}{nDCL}
\DeclareMathOperator*{\ndcg}{nDCG}
\DeclareMathOperator*{\dcl}{DCL}
\DeclareMathOperator*{\dcg}{DCG}
\DeclareMathOperator*{\idcl}{iDCL}
\DeclareMathOperator*{\idcg}{iDCG}

\DeclareGraphicsExtensions{.pdf, .pgf, .eps}
\graphicspath{ {figures/} }

\begin{document}

\CopyrightYear{2016}
\setcopyright{acmlicensed}
\conferenceinfo{RecSys '16,}{September 15--19, 2016, Boston, MA, USA.}
\isbn{978-1-4503-4035-9/16/09}\acmPrice{\$15.00}
\doi{http://dx.doi.org/10.1145/2959100.2959170}

\clubpenalty=10000
\widowpenalty=10000

\title{Fifty Shades of Ratings: How to Benefit from a Negative Feedback in Top-N Recommendations Tasks}

\numberofauthors{2}

\author{
\alignauthor Evgeny Frolov\\
    \email{evgeny.frolov@skoltech.ru}\\
    \affaddr{Skolkovo Institute of Science and Technology}\\
    \affaddr{143025, Nobel St. 3, Skolkovo Innovation Center}\\
    \affaddr{Moscow, Russia}
\alignauthor Ivan Oseledets\\
    \email{i.oseledets@skoltech.ru}\\
    \affaddr{Skolkovo Institute of Science and Technology}\\
    \affaddr{Institute of Numerical Mathematics of the Russian Academy of Sciences}\\
    \affaddr{119333, Gubkina St. 8}\\
    \affaddr{Moscow, Russia}
}

\maketitle
\begin{abstract}
Conventional collaborative filtering techniques treat a \mbox{top-$n$} recommendations problem as a task of generating a list of the most relevant items. This formulation, however, disregards an opposite -- avoiding recommendations with completely \mbox{irrelevant} items. Due to that bias, standard algorithms, as well as commonly used evaluation metrics, become insensitive to negative feedback.
In order to resolve this problem we propose to treat user feedback as a categorical variable and model it with users and items in a ternary way. We employ a third-order tensor factorization technique and implement a higher order folding-in method to support online recommendations. The method is equally sensitive to entire spectrum of user ratings and is able to accurately predict relevant items even from a negative only feedback. Our method may partially eliminate the need for complicated rating elicitation process as it provides means for personalized recommendations from the very beginning of an interaction with a recommender system. We also propose a modification of standard metrics which helps to reveal unwanted biases and account for sensitivity to a negative feedback. Our model achieves state-of-the-art quality in standard recommendation tasks while significantly outperforming other methods in the cold-start ``no-positive-feedback'' scenarios.
\end{abstract}


\section{Introduction}
\label{sec:intro}
%
One of the main challenges faced across different recommender systems is a \emph{cold-start} problem. For example, in a user cold-start scenario, when a new (or unrecognized) user is introduced to the system and no side information is available, it is impossible to generate relevant recommendations without asking the user to provide initial feedback on some items. Randomly picking items for this purpose might be ineffective and frustrating for the user. A more common approach, typically referred as a rating elicitation, is to provide a pre-calculated non-personalized list of the most representative items. However, making this list, that on the one hand helps to better learn the user preferences, and on the other hand does not lead to the user boredom, is a non-trivial task and still is a subject for active research.

The problem becomes even worse if a pre-built list of items resonate poorly with the user's tastes, resulting in mostly negative feedback (e.g.\ items that get low scores or low ratings from the user). Conventional collaborative filtering algorithms, such as matrix factorization or similarity-based models, tend to favor similar items, which are likely to be irrelevant in that case. This is typically avoided by generating more items, until enough positive feedback (e.g.\ items with high scores or high ratings) is collected and relevant recommendations can be made. However, this makes engagement with the system less effortless for the user and may lead to a loss of interest in it.

We argue that these problems can be alleviated if the system is able to learn equally well from both positive and negative feedback. Consider the following movie recommendation example: a new user marks the ``Scarface'' movie with a low rating, e.g.\ 2 stars out of 5, and no other information is present in his or her profile. This is likely to indicate that the user does not like movies about crime and violence.
It also seems natural to assume that the user probably prefers
``opposite'' features, such as sentimental story (which can be present in romantic movies or drama), or happy and joyful narrative (provided by animation or comedies). In this case, asking to rate or recommending the ``Godfather'' movie is definitely a redundant and inappropriate action.
Similarly, if a user provides some negative feedback for the first part of a series (e.g.\ the first movie from the ``Lord of the rings'' trilogy), it is quite natural to expect that the system will not immediately recommend another part from the same series.

A more proper way to engage with the user in that case is to
leverage a sort of \emph{``users, who dislike that item, do like these items instead''} scenario.
Users certainly can share preferences not only in what they like, but also in what they do not like and it is fair to expect that techniques, based on collaborative filtering approach, could exploit this for more accurate predictions even from a solely negative feedback. In addition to that, a negative feedback may have a greater importance for a user, than a positive one. Some psychological studies demonstrate, that not only emotionally negative experience has a stronger impact on an individual's memory \cite{kensinger:2009}, but also have a greater effect on humans behavior in general \cite{rozin:2001}, known as the \emph{negativity bias}.

Of course, a number of heuristics or tweaks could be proposed for traditional techniques to fix the problem, however there are intrinsic limitations within the models that make the task hardly solvable. For example, algorithms could start looking for less similar items in the presence of an item with a negative feedback. However, there is a problem of preserving relevance to the user's tastes. It is not enough to simply pick the most dissimilar items, as they are most likely to loose the connection to user preferences. Moreover, it is not even clear when to switch between the ``least similar'' and the ``most similar'' modes. If a user assigns a 3 star rating for a movie, does it mean that the system still has to look for the least similar items or should it switch back to the most similar ones?
User-based similarity approach is also problematic, as it tends to generate a very broad set of recommendations with a mix of similar and dissimilar items, which again leads to the problem of extracting the most relevant, yet unlike recommendations.

In order to deal with the denoted problems we propose a new tensor-based model, that treats feedback data as a special type of categorical variable. We show that our approach not only improves user cold-start scenarios, but also increases general recommendations accuracy. The contributions of this paper are three-fold:
\begin{itemize}
    \itemsep0em
    \item We introduce a collaborative filtering model based on a third order tensor factorization technique. In contrast to commonly used approach, our model treats explicit feedback (such as movie ratings) not as a cardinal, but as an ordinal utility measure. The model benefits greatly from such a representation and provides a much richer information about all possible user preferences. We call it \emph{shades of ratings}.

    \item We demonstrate that our model is equally sensitive to both positive and negative user feedback, which not only improves recommendations quality, but also reduces the efforts needed to learn user preferences in cold-start scenarios.

    \item We also propose a \emph{higher order folding-in} method for real-time generation of recommendations in support to cold-start scenarios.  The method does not require recomputation of the full tensor-based model for serving new users and can be used online.
\end{itemize}

\section{Problem formulation}
\label{sec:problem}
The goal for conventional recommender system is to be able to accurately generate a personalized list of new and interesting items (\emph{top-n recommendations}), given a sufficient number of examples with user preferences. As has been noted, if preferences are unknown this requires special techniques, such as rating elicitation, to be involved first. In order to avoid that extra step we introduce the following additional requirements for a recommender system:
\begin{itemize}
    \itemsep0em
    \item the system must be sensitive to a full user feedback scale and not disregard its negative part,
    \item the system must be able to respond properly even to a single feedback and take into account its type (positive or negative).
\end{itemize}

These requirements should help to gently navigate new users through the catalog of items, making the experience highly personalized, as after each new step the system narrows down user preferences.

\subsection{Limitations of traditional models}
\label{subsec:limitations}
Let us consider without the loss of generality the problem of movies recommendations. Traditionally, this is formulated as a prediction task:
\begin{equation}
    \label{eq:oldmodel}
    f_R: \mathrm{User} \times \mathrm{Movie} \rightarrow \mathrm{Rating},
\end{equation}
where User is a set of all users, Movie is a set of all movies and $f_R$ is a utility function, that assigns predicted values of ratings to every (\emph{user, movie}) pair. In collaborative filtering models the utility function is learned from a prior history of interactions, i.e.\ previous examples of how users rate movies, which can be conveniently represented in the form of a matrix $R$ with $M$ rows corresponding to the number of users and $N$ columns corresponding to the number of movies. Elements $r_{ij}$ of the matrix $R$ denote actual movie ratings assigned by users. As users tend to provide feedback only for a small set of movies, not all entries of $R$ are known, and the utility function is expected to infer the rest values.

In order to provide recommendations, the predicted values of ratings $\hat r_{ij}$ are used to rank movies and build a ranked list of top-$n$ recommendations, that in the simplest case is generated as:
\begin{equation}
    \label{eq:top-n}
    \text{toprec}(i, n) := \argmax_{j \in \mathrm{Movie}}^n \hat r_{ij}.
\end{equation}
where toprec($i, n$) is a list of $n$ top-ranked movies predicted for a user $i$. The way the values of $\hat r_{ij}$ are calculated depends on a collaborative filtering algorithm and we argue that standard algorithms are unable to accurately predict relevant movies given only an example of user preferences with low ratings.

\subsubsection{Matrix factorization}
Let us first start with a matrix factorization approach.
As we are not aiming to predict the exact values of ratings and more interested in correct ranking, it is adequate to employ the singular value decomposition (SVD) \cite{golub1970singular} for this task. Originally, SVD is not defined for data with missing values, however, in case of top-$n$ recommendations formulation one can safely impute missing entries with zeroes, especially taking into account that pure SVD can provide state-of-the-art quality \cite{Cremonesi2010, Lee2016a}. Note, that by SVD we mean its truncated form of rank $r < min(M, N)$, which can be expressed as:
\begin{equation}
    \label{eq:svd}
    R \approx U \Sigma V^T \equiv PQ^T,
\end{equation}
where $U \in \mathbb{R}^{M \times r}, V \in \mathbb{R}^{N \times r}$ are orthogonal factor matrices, that embed users and movies respectively onto a lower dimensional space of latent (or hidden) features, $\Sigma$ is a matrix of singular values $\sigma_1 \geq \ldots \geq \sigma_r > 0$ that define the strength or the contribution of every latent feature into resulting score. We also provide an equivalent form with factors $P = U \Sigma^{\frac{1}{2}}$ and $Q = V \Sigma^{\frac{1}{2}}$, commonly used in other matrix factorization techniques.

One of the key properties, that is provided by SVD and not by many other factorization techniques is an orthogonality of factors $U$ and $V$ right ``out of the box''. This property helps to find approximate values of ratings even for users that were not a part of original matrix $R$. Using a well known \emph{folding-in} approach \cite{ekstrand2011collaborative}, one can easily obtain an expression for predicted values of ratings:
\begin{equation}
    \label{eq:folding-in}
    \boldsymbol{r} \approx VV^T\boldsymbol{p},
\end{equation}
where $\boldsymbol{p}$ is a (sparse) vector of initial user preferences, i.e.\ a vector of lenght $M$, where a position of every non-zero element corresponds to a movie, rated by a new user, and its value corresponds to the actual user's feedback on that movie. Respectively, $\boldsymbol{r}$ is a (dense) vector of length $M$ of all predicted movie ratings. Note, that for an existing user (e.g.\ whos preferences are present in matrix $R$) it returns exactly the corresponding values of SVD.

Due to factors orthogonality the expression can be treated as a projection of user preferences onto a space of latent features. It also provides means for quick recommendations generation, as no recomputation of SVD is required, and thus is suitable for online engagement with new users.
It should be noted also, that the expression \eqref{eq:folding-in} does not hold if matrix $V$ is not orthogonal, which is typically the case in many other factorization techniques (and even for $Q$ matrix in \eqref{eq:svd}). However, it can be transformed to an orthogonal form with QR decomposition.  


Nevertheless, there is a subtle issue here. If, for instance, $\boldsymbol{p}$ contains only a single rating, then it does not matter what exact value it has. Different values of the rating will simply scale all the resulting scores, given by \eqref{eq:folding-in}, and will not affect the actual ranking of recommendations. In other words, if a user provides a single 2-star rating for some movie, the recommendations list is going to be the same, as if a 5-star rating for that movie is provided.

\subsubsection{Similarity-based approach}
\label{subsec:knn}
It may seem that the problem can be alleviated if we exploit some user similarity technique. Indeed, if users share not only what they like, but also what they dislike, then users, similar to the one with a negative only feedback, might give a good list of candidate movies for recommendations. The list can be generated with help of the following expression:
\begin{equation}
    \label{eq:knn}
    r_{ij} = \frac{1}{K} \sum_{k \in S_i} (r_{kj}) \ \text{sim}(i, k),
\end{equation}
where $S_i$ is a set of users, the most similar to user $i$, sim($i, k$) is some similarity measure
between users $i$ and $k$ and $K$ is a normalizing factor, equal to $\sum_{i \in S_i} \left| sim(i, k) \right|$ in the simplest case.
The similarity between users can be computed by comparing either their latent features (given by the matrix $U$) or simply the rows of initial matrix $R$, which turns the task into a traditional user-based \emph{k-Nearest Neighbors} problem (kNN).
It can be also modified to a more advanced forms, that take into account user biases \cite{Adomavicius2005a}.

However, even though more relevant items are likely to get higher scores in user-similarity approach, it still does not guarantee an isolation of irrelevant items. Let us demonstrate it on a simple example. For the illustration purposes we will use a simple kNN approach, based on a cosine similarity measure. However, it can be generalized to more advanced variations of \eqref{eq:knn}.

Let a new user Tom have rated the ``Scarface'' movie with rating 2 (see Table \ref{tab:ratings}) and we need to decide which of two other movies, namely ``Toy Story'' or ``Godfather'', should be recommended to Tom, given an information on how other users - Alice, Bob and Carol - have also rated these movies.

As it can be seen, Alice and Carol, similarly to Tom, do not like criminal movies. They also both enjoy the ``Toy Story'' animation. Even though Bob demonstrates an opposite set of interests, the preferences of Alice and Carol prevail. From here it can be concluded that the most relevant (or safe) recommendation for Tom would be the ``Toy Story''. Nevertheless, the prediction formula \eqref{eq:knn} assigns the highest score to the ``Godfather'' movie, which is a result of a higher value of cosine similarity between Bob's and Tom's preference vectors.
\begin{table}
\scriptsize
\centering
\caption{Similarity-based recommendations issue.}
\begin{tabular}{l c c c }
\multicolumn{1}{c}{} & Scarface & Toy Story & Godfather \\ \cline{2-4} 
\multicolumn{1}{c}{} & \multicolumn{3}{c}{\textit{Observation}} \\ \hline
Alice & 2 & 5 & 3 \\ 
Bob & 4 &   & 5 \\ 
Carol & 2 & 5 &  \\ 
\multicolumn{1}{c}{} & \multicolumn{3}{c}{\textit{New user}} \\ \hline
Tom & 2 & ? & ? \\
\multicolumn{2}{c}{} & \multicolumn{2}{c}{\textit{Prediction}} \\ \cline{3-4}
& & 2.6 & \textbf{3.1} \\
\end{tabular}
\label{tab:ratings}
\end{table}

\subsection{Resolving the inconsistencies}
\label{subsec:themodel}
The problems, described above, suggest that in order to build a model, that fulfills the requirements, proposed in Section \ref{sec:problem}, we have to move away from traditional representation of ratings.
Our idea is  to restate the problem formulation in the following way:
\begin{equation}
    \label{eq:newmodel}
    f_R: \mathrm{User} \times \mathrm{Movie} \times \mathrm{Rating} \rightarrow \mathrm{Relevance \ Score},
\end{equation}
where Rating is a domain of ordinal (categorical) variables, consisting of all possible user ratings, and Relevance Score denotes the likeliness of observing a certain (\emph{user, movie, rating}) triplet.
With this formulation relations between users, movies and ratings are modelled in a ternary way, i.e.\ all three variables influence each other and the resulting score.
This type of relations can be modelled with several methods, such as Factorization Machines \cite{rendle2011fast} or other context-aware methods \cite{Adomavicius2011}. We propose to solve the problem with a tensor-based approach, as it seems to be more flexible, naturally fit the formulation \eqref{eq:newmodel} and has a number of advantages, described in Section \ref{sec:tensors}. 

\begin{figure}[h]
\centering{
\includegraphics[width=65mm]{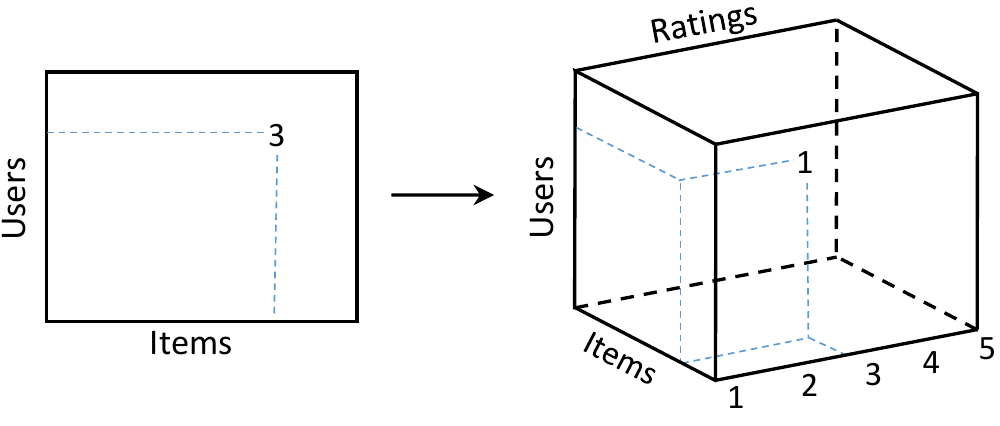}
}
\caption{From a matrix to a third order tensor.}
\label{fig:restate}
\end{figure}

\section{Tensor concepts}
\label{sec:tensors}
The (\emph{user, movie, rating}) triplets can be encoded within a three-dimensional array (see Figure \ref{fig:restate}) which we will call a third order tensor and denote with calligraphic capital letter $\mathcal{X}{\,\in\,}\mathbb{R}^{M \times N \times K}$.
The sizes of tensor dimensions (or \emph{modes})
correspond to the total number of unique users, movies and ratings. The values of the tensor $\mathcal{X}$ are binary (see \mbox{Figure \ref{fig:restate}}):
\begin{equation}
    \begin{cases}
        x_{ijk} = 1, &\text{ if } (i,j,k) \in S,\\
        x_{ijk} = 0, &\text{ otherwise},
    \end{cases}
\end{equation}
where $S$ is a history of known interactions, i.e.\ a set of observed (\emph{user, movie, rating}) triplets. Similarly to a matrix factorization case \eqref{eq:svd}, we are interested in finding such a tensor factorization that reveals some common patterns in the data and finds latent representation of users, movies and ratings.

There are two commonly used tensor factorization techniques, namely Candecomp/Parafac (CP) and Tucker Decomposition (TD) \cite{Kolda2009}. As we will show in Section \ref{subsec:ho-folding-in}, the use of TD is more advantageous, as it gives orthogonal factors, that can be used for quick recommendations computation, similarly to \eqref{eq:folding-in}. Moreover, an optimization task for CP, in contrast to TD, is ill-posed in general \cite{Silva}.

\subsection{Tensor factorization}
The tensor in TD format can be represented as follows:
\begin{equation}
    \label{eq:tucker}
    \mathcal{X} \approx  \mathcal{G} \times_1U\times_2V\times_3W,
\end{equation}
where $\times_n$ is an $n$-mode product, defined in \cite{Kolda2009}, $U \in \mathbb{R}^{M \times r_1}$, $V \in \mathbb{R}^{N \times r_2}$, $W \in \mathbb{R}^{K \times r_3}$ are orthogonal matrices, that represent embedding of the users, movies and ratings onto a reduced space of latent features, similarly to SVD case.

Tensor $\mathcal{G} \in \mathbb{R}^{r_1 \times r_2 \times r_3}$ is called the core of TD and a tuple of numbers ($r_1, r_2, r_3$) is called a \emph{multilinear rank} of the decomposition. The decomposition can be effectively computed with a higher order orthogonal iterations (HOOI) algorithm \cite{de2000best}.

\begin{figure}[b]
\centering{
\includegraphics[width=65mm]{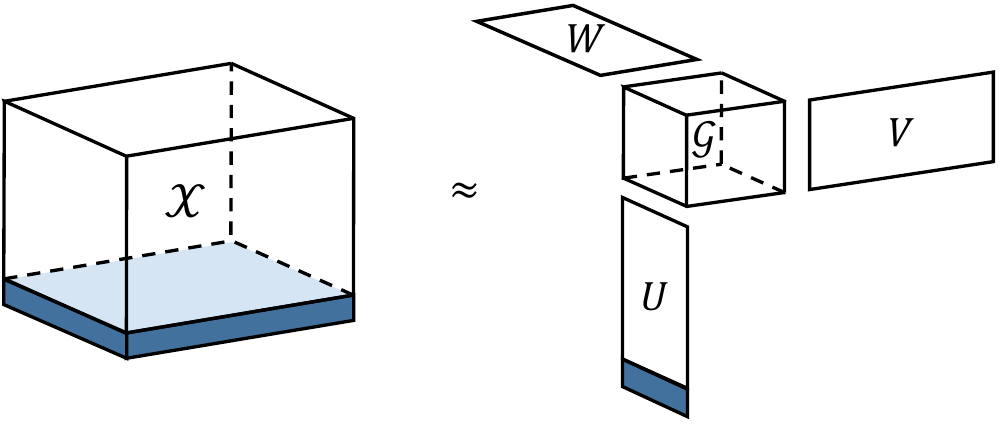}
}
\caption{Higher order folding-in for Tucker decomposition. A slice with a new user information in the original data (left) and a corresponding row update of the factor matrix in Tucker decomposition (right) are marked with solid a color.}
\label{fig:folding}
\end{figure}

\subsection{Efficient computation of recommendations}
\label{subsec:ho-folding-in}
As recommender systems typically have to deal with large numbers of users and items, this renders the problem of fast recommendations computation. Factorizing the tensor for every new user can take prohibitively long time which is inconsistent with the requirement of real-time recommendations. For this purposes we propose a \emph{higher order folding-in} method (see Figure \ref{fig:folding}) that finds approximate recommendations for any unseen user with comparatively low computational cost (cf.\ \eqref{eq:folding-in}):
\begin{equation}
    \label{eq:ho-folding-in}
    R_{(i)} = VV^TP_{(i)}WW^T,
\end{equation}
where $P_{(i)}$ is an $N \times K$ binary matrix of an $i$-th user's preferences and $R_{(i)} \in \mathbb{R}^{ N \times K}$ is a matrix of recommendations.
Similarly to SVD-based folding-in, \eqref{eq:ho-folding-in} can be treated as a sequence of projections to latent spaces of movies and ratings. Note, this is a straightforward generalization of matrix folding-in and we omit its derivation due to space limits. In the case of a known user the expression also gives the exact values of the TD.
%
%


\subsection{Shades of ratings}
\label{subsec:shades}
\begin{figure}
\centering{
\includegraphics[width=85mm,trim={0.1cm 0 0.1cm 0},clip]{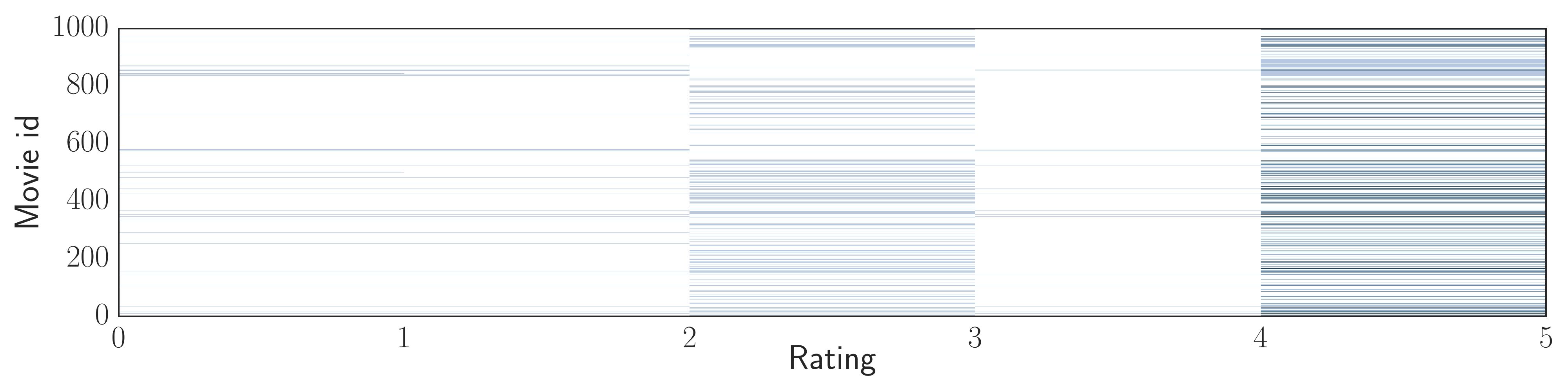}
}
\caption{Example of the predicted user preferences, that we call \emph{shades of ratings}. Every horizontal bar can be treated as a likeliness of some movie to have a specific rating for a selected user. More dense colors correspond to higher relevance scores. }
\label{fig:shades}
\end{figure}

Note that even though \eqref{eq:ho-folding-in} looks very similar to \eqref{eq:folding-in}, there is a substantial difference in what is being scored. In the case of a matrix factorization we score ratings (or other forms of feedback) directly, whereas in the tensor case we \emph{score the likeliness of a rating to have a particular value for an item}. This gives a new and more informative view on predicted user preferences (see Figure \ref{fig:shades}). Unlike the conventional methods, every movie in recommendations is not associated with just a single score, but rather with a full range of all possible rating values, that users are exposed to.
%



Another remarkable property of ``rating shades'' is that it can be naturally utilized for both ranking and rating prediction tasks. The ranking task corresponds to finding a maximum score along the movies mode (2nd mode of the tensor) for a selected (highest) rating. Note, that the ranking can be performed within every rating value. Rating prediction corresponds to a maximization of relevance scores along the ratings mode (i.e.\ the 3rd mode of the tensor) for a selected movie. We utilize this feature to additionally verify the model's quality (see Section \ref{sec:results}).

If a positive feedback is defined by several ratings (e.g.\ 5 and 4), than the sum of scores from these ratings can be used for ranking. Our experiments show that this typically leads to an improved quality of predictions comparing to an unmodified version of an algorithm.

\section{Evaluation}
\label{sec:evaluation}
As has been discussed in Section \ref{subsec:limitations}, standard recommender models are unable to properly operate with a negative feedback and more often simply ignore it. As an example, a well known recommender systems library MyMediaLite \cite{Gantner2011MyMediaLite}, that features many state-of-the-art algorithms, does not support a negative feedback for item recommendation tasks.

In addition to that, a common way of performing an offline evaluation of recommendations quality is to measure only how well a tested algorithm can retrieve highly relevant items. Nevertheless, both relevance-based (e.g.\ precision, recall, etc.) and ranking-based (e.g.\ nDCG, MAP, etc.) metrics, are completely insensitive to irrelevant items prediction: an algorithm that recommends 3 positively rated and 7 negatively rated items will gain the same evaluation score as an algorithm that recommends 3 positively rated and 7 items with unknown (not necessarily negative) ratings.

This leads to several important questions, that are typically obscured and that we aim to find an answer to:
\begin{itemize}
    \itemsep0em
    \item How likely an algorithm is to place irrelevant items in top-$n$ recommendations list and rank them highly?
    \item Does high evaluation performance in terms of relevant items prediction guarantee a lower number of irrelevant recommendations?
\end{itemize}
Answering these questions is impossible within standard evaluation paradigm and  we propose to adopt commonly used metrics in a way that respects crucial difference between the effects of relevant and irrelevant recommendations. We also expect that modified metrics will reflect the effects, described in Section \ref{sec:intro} (the Scarface and Godfather example).

\subsection{Negativity bias compliance}
\label{subsec:metrics}
The first step for the metrics modification is to split rating values into 2 classes: the class of a negative feedback and the class of a positive feedback. This is done by selecting a \emph{negativity threshold} value, such that the values of ratings above this threshold are treated as positive examples and all other values - as negative.

\begin{figure}
\centering{
\includegraphics[width=60mm,trim={0 1.9cm 0 0.1cm},clip]{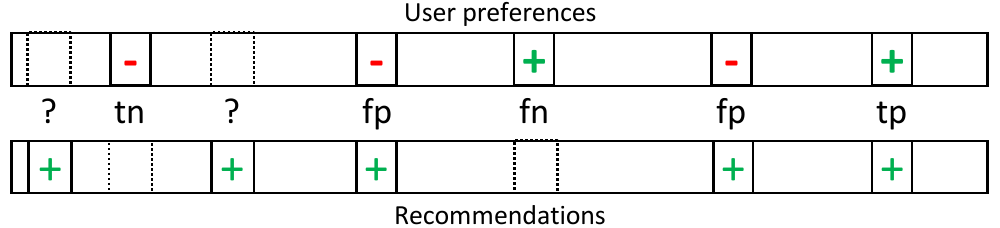}
}
\caption{Definition of matching and mismatching predictions. Recommendations that are not a part of the known user preferences (question marks) are ignored and not considered as false positive.
}
\label{fig:hits}
\end{figure}

The next step is to allow generated recommendations to be evaluated against the negative user feedback, as well as the positive one.
This leads to a classical notion of true positive (tp), true negative (tn), false positive (fp) and false negative (fn) types of predictions \cite{Shani2011}, which also renders a classical definition of relevance metrics, namely precision ($P$) and recall ($R$, also referred as \emph{True Positive Rate} (TPR)):
\begin{displaymath}
    P = \frac{tp}{tp + fp}, \quad R = \frac{tp}{tp + fn}.
\end{displaymath}
Similarly, \emph{False Positive Rate} (FPR) is defined as
\begin{displaymath}
    FPR = \frac{fp}{tp + fp}.
\end{displaymath}
The TPR to FPR curve, also known as a Receiver Operating Characteristics (ROC) curve, can be used to assess the tendency of an algorithm to recommend irrelevant items.
Worth noting here, that if items, recommended by an algorithm, are not rated by a user (question marks on Figure \ref{fig:hits}), then we simply ignore them and do not mark as false positive in order to avoid \emph{fp} rate overestimation \cite{Shani2011}.



The \emph{Discounted Cumulative Gain} (DCG) metric will look very similar to the original one with the exception that we do not include the negative ratings into the calculations at all:
\begin{equation}
    \label{eq:dcg}
    \dcg = \sum_{p} \frac{2^{r_p} - 1}{\log_2{(p+1)}},
\end{equation}
where $p: \{r_p > \text{negativity threshold}\}$ and $r_p$ is a rating of a positively rated item. This gives an nDCG metric:
\begin{displaymath}
    \ndcg = \frac{\dcg}{\idcg},
\end{displaymath}
where $\idcg$ is a value returned by an ideal ordering or recommended items (i.e.\ when more relevant items are ranked higher in top-$n$ recommendations list).

\subsection{Penalizing irrelevant recommendations}
The nDCG metric indicates how close tp predictions are to the beginning of a top-$n$ recommendations list, however, it tells nothing about the ranking of irrelevant items. We fix this by a modification of \eqref{eq:dcg} with respect to a negative feedback, which we call a \emph{Discounted Cumulative Loss}:
\begin{equation}
    \label{eq:dcl}
    \dcl = \sum_{n} \frac{2^{-r_n} - 1}{-\log_2{(n+1)}},
\end{equation}
where $n: \{r_n \leq \text{negativity threshold}\}$ and $r_n$ is a rating of a negatively rated item. Similarly to nDCG, nDCL metric is defined as:
\begin{displaymath}
    \ndcl = \frac{\dcl}{\idcl},
\end{displaymath}
where $\idcl$ is a value returned by an ideal ranking or irrelevant predictions (i.e.\ the more irrelevant are ranked lower). Note, that as $\ndcl$ penalizes high ranking of irrelevant items, therefore \emph{the lower are the values of nDCL the better}.

In the experiments all the metrics are measured for different values of top-$n$ list length, i.e.\ the metrics are metrics \emph{at n}. The values of metrics are averaged over all test users.



\subsection{Evaluation methodology}
\label{subsec:evalmeth}
For the evaluation purposes we split datasets into two subsets, disjoint by users (e.g.\ every user can only belong to a single subset). First subset is used for learning a model, it contains 80\% of all users and is called a \emph{training set}.  The remaining 20\% of users (the test users) are unseen in the training set and are used for models evaluation. We holdout a fixed number of items from every test user and put them into a \emph{holdout set}. The remaining items form an \emph{observation set} of the test users. Recommendations, generated based on an observation set are evaluated against a holdout set. We also perform a 5-fold cross validation by selecting different 20\% of users each time and averaging the results over all 5 folds.
The main difference from common evaluation methodologies is that we allow both relevant and irrelevant items in the holdout set (see Figure \ref{fig:hits} and Section \ref{subsec:metrics}).
Furthermore, we vary the number and the type of items in the observation set, which leads to various scenarios:
\begin{itemize}
\itemsep0mm
    \item leaving only one or few items with the lowest ratings leads to the case of ``no-positive-feedback'' cold-start;
    \item if all the observation set items are used to predict user preferences, this serves as a proxy to a standard recommendation scenario for a known user.
\end{itemize}

We have also conducted rating prediction experiments when a single top-rated item is held out from every test user (see Section \ref{sec:results}). In this experiment we verify what ratings are predicted by our model (see Section \ref{subsec:shades} for explanation of rating calculation) against the actual ratings of the holdout items.


\section{Experimental setup}
\label{sec:experiments}

\subsection{Datasets}
We use publicly available Movielens\footnote{https://grouplens.org/datasets/movielens/} 1M and 10M datasets as a common standard for offline recommender systems evaluation. We have also trained a few models on the latest Movielens dataset (22M rating, updated on 1/2016) and deployed a movie recommendations web application for online evaluation.
This is especially handy for our specific scenarios, as the content of each movie is easily understood and contradictions in recommendations can be easily eye spotted (see Table \ref{tab:examples}).

\subsection{Algorithms}
We compare our approach with the state-of-the-art matrix factorization methods, designed for items recommendations task as well as two non-personalized baselines.

\begin{compactitem}
\item \emph{CoFFee} (Collaborative Full Feedback model) is the proposed tensor-based approach.
\end{compactitem}
\begin{compactitem}
\item  \emph{SVD}, also referred as \emph{PureSVD} \cite{Cremonesi2010}, uses standard SVD. As in the CoFFee model, missing values are simply imputed with zeros.
\end{compactitem}
\begin{compactitem}
\item \emph{WRMF} \cite{pan2008one} is a matrix factorization method, where missing ratings are not ignored or imputed with zeroes, but rather are uniformly weighted. 
\end{compactitem}
\begin{compactitem}
    \item \emph{BPRMF} \cite{Rendle2009a} is a matrix factorization method, powered by a Bayesian personalized ranking approach, that optimizes pair-wise preferences between observed and unobserved items. 
\end{compactitem}
\begin{compactitem}
\item \emph{Most popular} model always recommends \mbox{top-$n$} items with the highest number of ratings (independently of ratings value).
\end{compactitem}
\begin{compactitem}
\item \emph{Random guess} model generates recommendations randomly.
\end{compactitem}

SVD is based on Python's Numpy, and SciPy packages, which heavily use BLAS and LAPACK functions as well as MKL optimizations. CoFFee is also implemented in Python, using the same packages for most of the linear algebra operations. We additionally use Pandas package to support sparse tensors in COO format.

\begin{figure}
\centering{
\includegraphics[width=50mm,trim={0 0.7cm 0 0},clip]{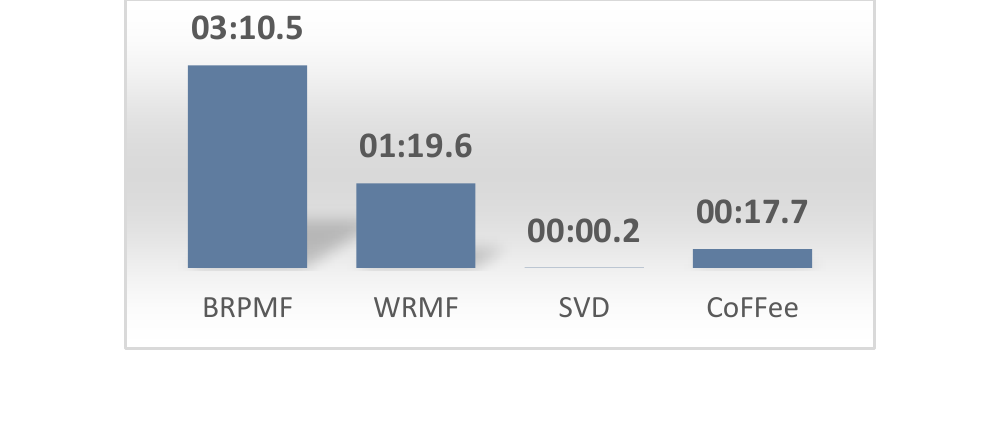}
}
\caption{Models' learning time, mm:ss.ms (single laptop, Intel i5 2.5GHz CPU, Movielens 10M).}
\label{fig:runtime}
\end{figure}

BPRMF and WRMF implementations are taken from the MyMediaLite \cite{Gantner2011MyMediaLite} library. We wrap the command line utilities of these methods with Python, so that all the tested algorithms share the same namespace and configuration. Command line utilities do not support online evaluation and we implement our own (orthogonalized) folding-in on the factor matrices generated by these models. Learning times of the models are depicted on Figure \ref{fig:runtime}. The source code as well as the link to our web app can be found at Github\footnote{https://github.com/Evfro/fifty-shades}.



\subsection{Settings}
We preprocess these datasets to contain only users who have rated no less than 20 movies. Number of holdout items is always set to 10. The top-$n$ values range from 1 to 100. The test set items selection is 1 or 3 negatively rated, 1, 3 or 5 with random ratings and all (see Section \ref{subsec:evalmeth} for details).
We include higher values of top-$n$ (up to 100) as we allow random items to appear in the holdout set. This helps to make experimentation more sensitive to wrong recommendations, that match negative feedback from the user.
The negativity threshold is set to 3 for Movielens 1M and 3.5 for Movielens 10M. Both observation and holdout sets are cleaned from the items that are not present in the training set. The number of latent factors for all matrix factorization models is set to 10, CoFFee multilinear rank is (13, 10, 2). Regularization parameters of WRMF and BPRMF algorithms are set to MyMediaLite's defaults.






\section{Results}
\label{sec:results}

\begin{table*}
\scriptsize
\centering
\caption{Hand-picked examples from top-10 recommendations generated on a single feedback. The models are trained on the latest Movielens dataset.}
\begin{tabular}{l c c c }
    &
    \begin{tabular}{c c}
    Scarface \\ \centering{\includegraphics[width=12mm,trim={0 2.5cm 0 0},clip]{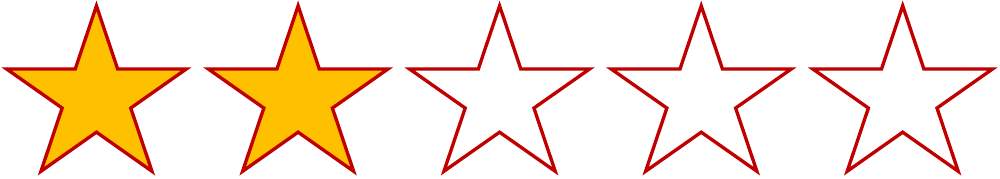}}
    \end{tabular}
    &
    \begin{tabular}{c c}
    {LOTR: The Two Towers} \\ \centering{\includegraphics[width=12mm,trim={0 2.5cm 0 0},clip]{figures/2stars.pdf}}
    \end{tabular}
    &
    \begin{tabular}{c c}
    Star Wars: Episode VII - The Force Awakens \\ \centering{\includegraphics[width=12mm,trim={0 2.5cm 0 0},clip]{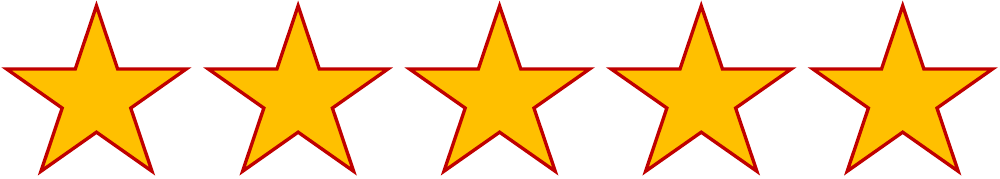}}
    \end{tabular} \\ \hline \hline
     \multirow{3}{*}{CoFFee} & Toy Story & Net, The & Dark Knight, The \\
     & Mr. Holland's Opus & Cliffhanger & Batman Begins\\
     & Independence Day & Batman Forever & Star Wars: Episode IV - A New Hope\\ \hline
     \multirow{3}{*}{SVD} & Reservoir Dogs & LOTR: The Fellowship of the Ring & Dark Knight, The \\
     & Goodfellas & Shrek & Inception \\
     & Godfather: Part II, The & LOTR: The Return of the King & Iron Man\\ \hline
\end{tabular}
\label{tab:examples}
\end{table*}

\begin{figure*}[!thb]
\centering{
\includegraphics[width=176mm]{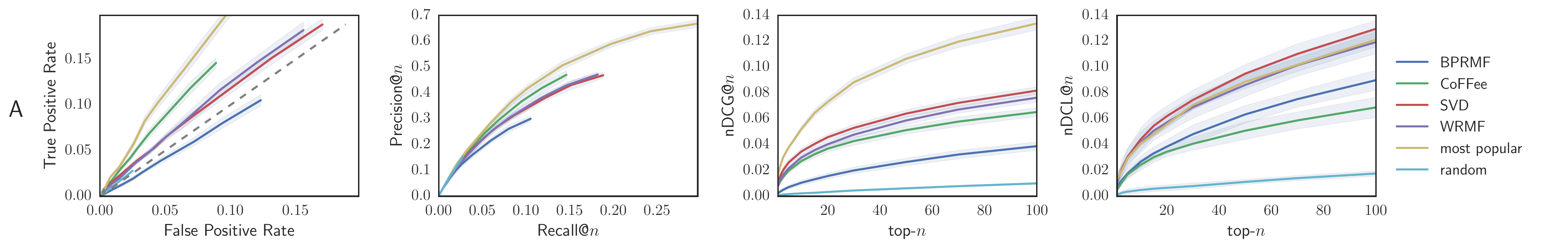}
\includegraphics[width=176mm]{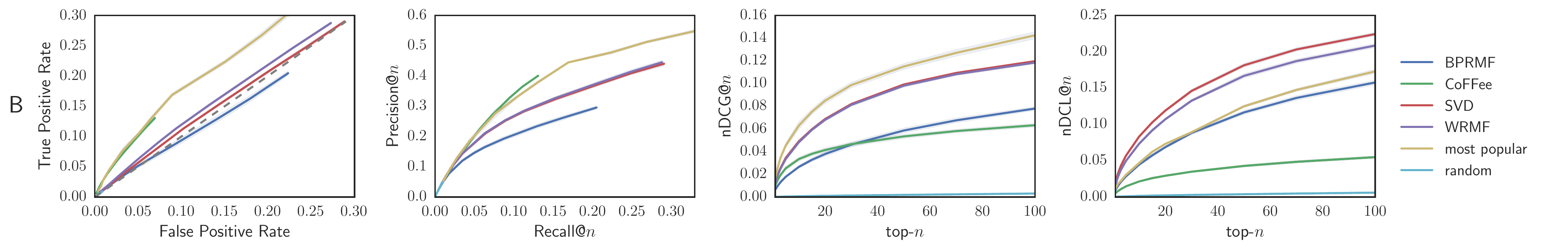}
\includegraphics[width=175mm]{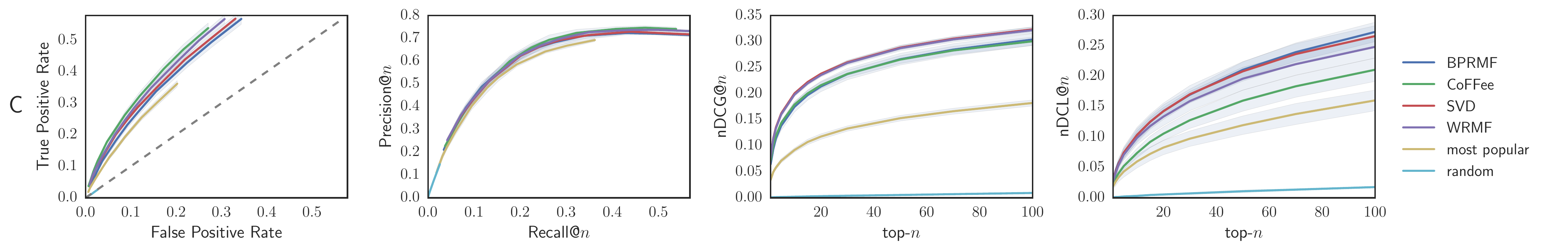}
\includegraphics[width=175mm]{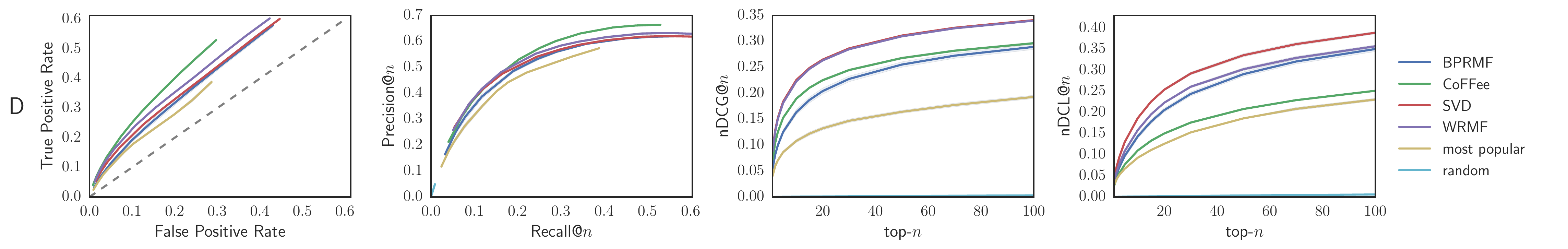}
}
\caption{The ROC curves (1st column), precision-recall curves (2nd column), nDCG@$n$ (3rd column) and nDCL@$n$ (4th column). Rows A, B correspond to a cold-start with a \emph{single negative} feedback. Rows C, D correspond to a \emph{known user} recommendation scenario. Odd rows are for Movielens 1M, even rows are for Movielens 10M. For the first 3 columns the higher the curve, the better, for the last column the lower the curve, the better. Shaded areas show a standard deviation of an averaged over all cross validation runs value.}
\label{fig:results}
\end{figure*}


Due to a space constraints we provide only the most important and informative part of evaluation results. They are presented on the Figure \ref{fig:results}.  Rows A and C correspond to Movielens 1M dataset, rows B and D correspond to Movielens 10M dataset. We also report a few interesting hand-picked examples of movies recommendations, generated from a single negative feedback (see Table \ref{tab:examples}).

\paragraph{How to read the graphs}
The results are better understood with particular examples. Let us start with the first two rows on Figure \ref{fig:results} (row A is for Movielens 1M and row B is for Movielens 10M). These rows correspond to a performance of the models, when only a single (random) negative feedback is provided.

First of all, it can be seen that the item popularity model gets very high scores with TPR to FPR, precision-recall and $\ndcg$ metrics (first 3 columns on the figure). One could conclude that this is the most appropriate model in that case (when almost nothing is know about user preferences). However, high $\ndcl$ score (4th column) of this model indicates that it is also very likely to place irrelevant items at the first place, which can be disappointing for users. Similar poor ranking of irrelevant items is observed with SVD and WRMF models. On the other hand, the lowest $\ndcl$ score belongs to the random guess model, which is trivially due to a very poor overall performance. The same conclusion is valid for BPRMF model, that have low $\ndcl$ (row A), but fails to recommend relevant items from a negative feedback.

The only model, that behaves reasonably well is the proposed CoFFee model. It has low $\ndcl$, i.e.\ \emph{it is more successful at avoiding irrelevant recommendations at the first place}. This effect is especially strong on the Movielens 10M dataset (row B). The model also exhibits a better or comparable to the item popularity model's performance on relevant items prediction. At first glance, the surprising fact is that the model has a low $\ndcg$ score. Considering the fact that this can not be due to a higher ranking of irrelevant items (as it follows from low $\ndcl$), this is simply due to a higher ranking of items, that were not yet rated by a user (recall the question marks on Figure \ref{fig:hits}).

The model tries to make a \emph{safe guess} by filtering out irrelevant items and proposing those items that are more likely to be relevant to an original negative feedback (unlike popular or similar items recommendation). This conclusion is also supported by the examples from the first 2 columns of Table \ref{tab:examples}. It can be easily seen, that unlike SVD, the CoFFee model makes safe recommendations with ``opposite'' movie features (e.g.\ Toy Story against Scarface). Such an effects are not captured by standard metrics and can be revealed only by a side by side comparison with the proposed $\ndcl$ measure.

In standard recommendations scenario, when user preferences are known (rows C, D) our model also demonstrates the best performance in all but $\ndcg$ metrics, which again is explained by the presence of unrated items rather than a poor quality of recommendations. In contrast, matrix factorization models, SVD and WRMF, while also being the top-performers in the first three metrics, demonstrate the worst quality in terms of  $\ndcl$ almost in all cases.

We additionally test our model in rating prediction experiment, where the ratings of the holdout items are predicted as described in Section \ref{subsec:shades}. On the Movielens 1M dataset our model is able to predict the exact rating value in 47\% cases. It also correctly predicts the rating positivity (e.g.\ predicts rating 4 when actual rating is 5 and vice versa) in another 48\% of cases, giving 95\% of correctly predicted feedback positivity in total. As a result it achieves a 0.77 RMSE score on the holdout set for Movielens 1M.


\section{Related work}
A few research papers have studied an effect of different types of user feedback on the quality of recommender systems.
The authors of Preference model \cite{Lee2016a} proposed to split user ratings into categories and compute the relevance scores based on user-specific ratings distribution.
The authors of SLIM method \cite{ning2011slim} have compared models that learn ratings either explicitly (r-models) or in a binary form (b-models). They compare initial distribution of ratings with the recommendations by calculating a hit rate for every rating value. The authors show that r-models have a stronger ability to predict top-rated items even if ratings with the highest values are not prevalent.
The authors of \cite{Amatriain:2009} have studied the subjective nature of ratings from a user perspective. They have demonstrated that a rating scale per se is non-uniform, e.g.\ distances between different rating values are perceived differently even by the same user.
The authors of \cite{chao2005adaptive} state that discovering what a user does not like can be easier than discovering what the user does like. They propose to filter all negative preferences of individuals to avoid unsatisfactory recommendations to a group.
However, to the best of our knowledge, there is no published research on the \emph{learning from an explicit negative feedback} paradigm in personalized recommendations task.

\section{Conclusion and perspectives}
To conclude, let us first address the two questions, posed in the beginning of Section \ref{sec:evaluation}. As we have shown, standard evaluation metrics, that do not treat irrelevant recommendations properly (as in the case with $\ndcg$), might obscure a significant part of a model's performance. An algorithm, that highly scores both relevant and irrelevant items, is more likely to be favored by such a metrics, while increasing the risk of a user disappointment.

We have proposed modifications to both standard metrics and evaluation procedure, that not only reveal a positivity bias of standard evaluation, but also help to perform a comprehensive examination of recommendations quality from the perspective of both positive and negative effects.

We have also proposed a new model, that is able to learn equally well from full spectrum of user feedbacks and provides state-of-the-art quality in different recommendation scenarios.
The model is unified in a sense, that it can be used both at initial step of learning user preferences and at standard recommendation scenarios for already known users. We believe that the model can be used to complement or even replace standard rating elicitation procedures and help to safely introduce new users to a recommender system, providing a highly personalized experience from the very beginning.





\section{Acknowledgements}
This material is based on work supported by the
Russian Science Foundation under grant 14-11-00659-a.

\bibliographystyle{abbrv}
\bibliography{literature}
\end{document}